\documentclass[conference]{IEEEtran}

\usepackage{multirow}
\usepackage{caption} 
\usepackage{subcaption}
\usepackage{amsfonts}

\usepackage{graphicx}% http://ctan.org/pkg/graphicx

\begin{document}

\title{A Multi-Modal Graph-Based Semi-Supervised Pipeline for Predicting Cancer Survival}

\author{\IEEEauthorblockN{Hamid Reza Hassanzadeh}
	\IEEEauthorblockA{Department of Computational Science\\ and Engineering\\Georgia Institute of Technology\\Atlanta, Georgia 30332\\
		Email: hassanzadeh@gatech.edu}
	\and
	\IEEEauthorblockN{John H. Phan}
	\IEEEauthorblockA{Department of Biomedical Engineering\\Georgia Institute of Technology\\and Emory University\\Atlanta, Georgia 30332\\
		Email: jhphan@gatech.edu}
	\and
	\IEEEauthorblockN{May D. Wang}
	\IEEEauthorblockA{Department of Biomedical Engineering\\Georgia Institute of Technology\\and Emory University\\Atlanta, Georgia 30332\\
		Email: maywang@bme.gatech.edu}
	
}

%For arXive.org
\newcommand{\MYfooter}{\smash{
		\hfil\parbox[t][\height][t]{\textwidth}{\centering
			\thepage}\hfil\hbox{}}}

\makeatletter
\def\ps@IEEEtitlepagestyle{%
	\def\@oddhead{\parbox[t][\height][t]{\textwidth}{%\centering
			Published in 2016 IEEE International Conference on Bioinformatics and Biomedicine (BIBM)
		}\hfil\hbox{}}%
	}

% make the title area

\maketitle

\begin{abstract}
Cancer survival prediction is an active area of research that can help prevent unnecessary therapies and improve patient's quality of life. Gene expression profiling is being widely used in cancer studies to discover informative biomarkers that aid predict different clinical endpoint prediction. We use multiple modalities of data derived from RNA deep-sequencing (RNA-seq) to predict survival of cancer patients. Despite the wealth of information available in expression profiles of cancer tumors, fulfilling the aforementioned objective remains a big challenge, for the most part, due to the paucity of data samples compared to the high dimension of the expression profiles. As such, analysis of transcriptomic data modalities calls for state-of-the-art big-data analytics techniques that can maximally use all the available data to discover the relevant information hidden within a significant amount of noise. In this paper, we propose a pipeline that predicts cancer patients' survival by exploiting the structure of the input (manifold learning) and by leveraging the unlabeled samples using Laplacian support vector machines, a graph-based semi supervised learning (GSSL) paradigm. We show that under certain circumstances, no single modality per se will result in the best accuracy and by fusing different models together via a stacked generalization strategy, we may boost the accuracy synergistically. We apply our approach to two cancer datasets and present promising results. We maintain that a similar pipeline can be used for predictive tasks where labeled samples are expensive to acquire.
\end{abstract}

%\keywords{Cancer survival prediction; Manifold learning; Graph-based semi-supervised learning; Laplacian support vector machines.}

\section{Introduction}\label{intro}
Cancer survival prediction provides valuable information for oncologists to choose an effective course of treatment for cancer patients. This is particularly important for terminally ill patients for whom uninformed treatment selection can undermine patients' quality of life \cite{paper1}. Unfortunately, there is a high amount of stochasticity involved in regard to this clinical endpoint due to numerous confounding factors (e.g., environmental factors) that can play a role in patients' survival. As a result, physicians often make subjective decisions based on their past experience. Yet another major challenge when dealing with clinical and omics data pertaining to the cancer patients is that a significant portion of these data are unlabeled due to the lack of adequate follow-up information. This is especially the case for the longitudinal studies that span several years (often more than five years,) which results in patient withdrawal from the study or are subject to research budget restrictions. As a result, most research studies filter out these samples and solve the problem within the framework of supervised learning. Even though this strategy works for larger datasets, it fails to produce accurate and robust results in most cancer datasets where small sample sizes remain a bottleneck. This, along with the high dimensionality of the input data, makes the objective difficult to achieve. An alternative to the above-mentioned strategy would be to leverage the information present in the available unlabeled data which are sometimes orders of magnitude bigger in size, compared to the amount of labeled samples. This approach falls into the category of the semi-supervised learning paradigm.

Semi-supervised learning (SSL) is a learning paradigm that takes advantage of both labeled and unlabeled data when training a model. Traditionally, models trained for predictive tasks are either supervised or unsupervised. While the labeled data convey valuable information that can be leveraged to train a model, they are not always available in abundance. On the other hand, the unlabeled data may be readily available in large quantities as they may be cheaper or faster to generate. In the case of cancer data, for example, it is often costly to conduct a long-term research study where the patient eventually dies due to his/her disease. Consequently, the outcome corresponding to the endpoint of study may not be known for a significant part of the dataset. Even though the unlabeled data do not tell us as much about the prediction outcome, may provide some information about the distribution of data that can be exploited to guide model training and lead us to a higher predictive power. However, the key requirement for any SSL method to succeed is that the marginal data distribution has to carry some useful information regarding the inference of the posterior probability \cite{paper2} (the Bayes rule.) 

To date, several SSL classification paradigms have been proposed (such as self-training, co-training, transductive SVM [TSVM] and graph-based semi-supervised learning [GSSL]). Despite the success of SSL in different domains, their application to the prediction of clinical endpoints is relatively new. Here, we give an account of the related published studies in chronological order. 

As one of the earliest applications of GSSL to tumor classification, Gui et al. \cite{paper3} used a graph-based approach named local and global consistency (LGC) \cite{paper4}. LGC is an iterative algorithm inspired from the spreading of activation networks from experimental psychology that can be explained as random walks on graphs. The main idea in LGC is that each unlabeled sample receives some amount of information from its neighbors, which is retained according to some learning rate. In the end, based on the amount of information each unlabeled node receives from other nodes, the label of that node is specified. Note that, one key difference between Laplacian SVM and LGC is that the latter is not considered to be a large margin classifier. Therefore, caution should be practiced when generalization is a concern. Shi et al. \cite{paper5} used transductive support vector machines (TSVM) to predict recurrence in colorectal cancer patients using microarray gene expression data. TSVM decides the labels for unlabeled samples such that the classification boundary can be placed in a low density region with the margin maximized. This approach, however, is not an appropriate choice when there is no clear boundary between the samples of different classes, which is often the case for cancer data. A similar approach \cite{paper6} for prediction of cancer subtypes has also been reported. In another successful application of GSSL, Kim et al. \cite{paper7} used manifold regularization to enforce smoothness constraints on the intrinsic geometry of the marginal input distribution. They applied their developed method to predict different clinical endpoints such as survival. Moreover, they used different modalities of genomic data and represented each with a graph Laplacian and showed that, by generating a graph Laplacian as a weighted sum of each individual Laplacian, the solution follows the same analytic form. However, due to the quadratic form of the cost function, it could be considered more of a semi-supervised extension to the regularized least squares method and lacks margin generalization. They used a similar approach in their follow-up studies \cite{paper8,paper9}. In another GSSL-related study, Kim et al. \cite{paper10} used two levels of semi-supervised learning to integrate different layers of omics data. Iteratively, they trained separate GSSL methods on each omics dataset, then used co-training to assign pseudo-labels to the unlabeled samples. More recent published research using similar techniques can be found in \cite{paper11,paper12,paper13,paper14,paper15}.

In this article, we develop a pipeline for prediction of cancer survival based on Laplacian support vector machines, an extension of support vector machines to the semi-supervised domain. This method incorporates the geometry of the marginal input distribution and maintains a balance between the loss function and semi-supervised smoothness assumptions where intuitively smoothness means data samples that are closer to each other in the input space, should also be closer in terms of the labels they are assigned.  Moreover, we compare and contrast the efficacy of the proposed method as a function of the size of the labeled and unlabeled sets.

\section{Manifold regularization \& Laplacian SVM}\label{background}
Semi-supervised learning differs from supervised learning in that the underlying dataset consists of a set of $l$ labeled inputs $(x_i,y_i)_{i=1}^l$ and a set of $u$ unlabeled samples $(x_i)_{i=l+1}^{l+u}$. As such, there is a probability distribution $P$ on $X\times \mathbb{R}$ from which examples are drawn. On the other hand, assuming that the unlabeled samples are drawn from the same underlying stochastic process, they should be distributed according to the marginal distribution $P_X$ of the original $P$. Graph-based semi-supervised learning (GSSL) \cite{paper4}, on the other hand, is based on the assumption that if the data lie on a manifold of much lower dimensions (the intrinsic geometry of $P_X$) than that of the input space, then by enforcing smoothness constraints on that manifold for both labeled and unlabeled samples, we can produce a classifier that is more consistent with the input data \cite{Chap06}. GSSL represents each data sample as a node in a weighted graph where the weight attributed to each edge, $w_{ij}$, is computed according to an affinity function, defined shortly. This results in a dense graph which is often not desired and hence, in a later step, sparsified using techniques such as \textit {k}-nearest neighbor. The resulting graph serves as a proxy for the manifold and is used within the framework of a variety of manifold regularization techniques. One important matrix that is defined over this graph and which plays an important role in derivation of many of the GSSL based approaches is the graph Laplacian, computed as $L=W-D$ where $W=[w_{ij}]_{i,j}$ is the affinity matrix and the diagonal matrix $D$ is given by $D_{ii}=\sum_{j=1}^{l+u}w_{ij}$. 

Proposed by Belkin et al. \cite{paper16}, manifold regularization is a family of semi-supervised learning algorithms that exploits the intrinsic geometry of the marginal distribution $P_X$ by adding an additional regularization term. The support of $P_X$ is assumed to have the geometric structure of a Riemannian manifold $M$ \cite{paper17}. Within this framework, one can derive an extension to the support vector machine by using the soft margin loss function and thereby incorporating the manifold structure of the input. As a result, the extended classifier not only leverages the manifold structure of the input but also remains a large margin classifier with useful generalization implications that make it an effective tool when dealing with data scarcity. Formally speaking, for any given Mercer kernel \cite{paper18},  $K:X\times X\to \mathbb{R}$, and the corresponding norm, $\Vert\,\Vert_K$, a number of popular algorithms including SVM can be cast into the following minimization form with different empirical cost functions:
\begin{equation}\label{eq1}
f^*=arg \min_{f\in \mathcal{H}_K } \frac{1}{l}\sum_{i=1}^l V(x_i,y_i,f)+\gamma \Vert f\Vert^2_\mathcal{H} 
\end{equation}
where $V$ is some loss function and $\mathcal{H}_K$ is the reproducing kernel Hilbert space (RKHS) of functions $X\to \mathbb{R}$ corresponding to kernel $K$. The regularization term in equation \ref{eq1} imposes smoothness conditions on possible solutions. According to the represented theorem the solution to equation \ref{eq1} exists in $\mathcal{H}_K$ and can be expressed as,
\begin{equation}\label{eq2}
f^* (x)=\sum_{i=1}^l\alpha_i K(x_i,x)
\end{equation}
and, as a result, the problem reduces to optimizing over the finite dimensional space of coefficient $\alpha_i$. The problem with this framework, however, is that it only enforces the smoothness constraint in the kernel space (aka the ambient space) but does not take into account the smoothness of the solution over the manifold, that is, the geometric structure of $P_X$. To achieve this, they add an extra regularizer to enforce the smoothness of the solution relative to the manifold which expands the minimization problem in equation \ref{eq1}, to the following:
\begin{equation}\label{eq3}
f^*=arg\min_{f\in \mathcal{H}_K}\frac{1}{l} \sum_{i=1}^l V(x_i,y_i,f)+\gamma_A \Vert f\Vert_{\mathcal{H}}^2+\gamma_I \Vert f\Vert_I^2
\end{equation}
where $\Vert f\Vert_I^2$ is the norm in the intrinsic space. The regularizer parameters $\gamma_A$, $\gamma_I$ control the smoothness of the solution relative to the ambient and the intrinsic spaces, respectively. It can be shown that when the intrinsic space is a compact manifold $M\in \mathbb{R}^n$ then one natural choice for $\Vert f\Vert_I^2$ would be
\begin {equation} \label{eq4}
\Vert f\Vert_I^2=\int_M \Vert \nabla_M f(x)\Vert ^2 dP_X  
\end{equation}
where $\nabla_M$ is the gradient of the function over the Riemannian manifold M and it can be approximated with the graph Laplacian corresponding to the proxy graph when the affinity matrix takes the exponential form $w_{ij}=e^{(-\frac{\Vert x_i-x_j \Vert}{\epsilon})}\; \forall i\ne j$ and $w_{ii}=0 \;\forall i$. This approximation reduces the intrinsic penalty term to 
\begin{equation} \label{eq5}
\Vert f \Vert_I^2=\mathbf {f}^T L\mathbf{f }
\end{equation}
where $\mathbf{f}^T=[f(x_1),\dots,f(x_{l+u} )]$. Finally, following from the representer theorem \cite{paper18}, they showed that the solution to equation \ref{eq3} with the soft-margin loss function and equation \ref{eq5} as the smoothness penalty term on the manifold (henceforth, Laplacian support vector machine) takes the same form as in equation \ref{eq2} but with a different set of weights $\alpha_i$.\\
\section{Materials and methods}
\subsection{Data}
We used RNA-seq data corresponding to cancer tissues for two cancer types, the kidney cancer (KIRC) and the neuroblastoma (NB) which occurs most often in infants and young children. The data for the former cancer was retrieved from the Cancer Genome Atlas (TCGA) data portal (http://www.tcga-data.nci.nih.gov) and for the latter we used the data from a previous published study \cite{Zhang15}. These data were generated using the Illumina HiSeq 2000 platform. In this study, we use three different feature levels (gene, transcript and junction) derived from these sequencing data. Table \ref{tab1} shows the data description for each modality and disease. We divided samples into positive and negative classes according to a selected patient survival threshold. For kidney cancer, this threshold is 5 years and for neuroblastoma cancer it is 9 years. 
\begin{table*}[]
\centering
\caption{Data description}
\label{tab1}
\begin{tabular}{|cc|c|c|}
Cancer type           & Platform          & Modality & \#Features \\
\hline
\multirow{4}{*}{NB} &    \multirow{8}{*}{Illumina HiSeq 2000}    & Isoform  & 263547\\
                      &                   & Gene     & 60781\\
                      &                   & Junction & 340417 \\\cline{3-4}
\multirow{4}{*}{KIRC} & & Isoform  & 73599      \\
                      &                   & Gene     & 20531      \\
                      &                   & Junction & 249567     
\end{tabular}
\end{table*}
Table \ref{tab2} illustrates the statistics of each class and the percentage of unlabeled samples. According to the table, more than half of the samples from each dataset are unlabeled which justifies the use of semi-supervised learning to benefit from them.

\begin{table*}[]
\centering
\caption{Number of labeled vs. unlabeled samples}
\label{tab2}
\begin{tabular}{ccccc}
\multirow{2}{*}{Cancer type}                                    & \multicolumn{2}{c}{\#Labeled Samples} & \#Unlabeled Samples & \%Unlabeled \\ \cline{2-3}
                                                                & \#Positive             & \#Negative             &        &       \\\hline
NB (9-y survival)                                           & 114                & 105                & 279     & 56\% \\\hline
KIRC (5-y survival)                                             & 110               & 140               & 278    & 53\%  
\end{tabular}
\end{table*}

\subsection{The proposed pipeline}
\begin{figure}[t!]
\centerline{
\includegraphics[scale=0.5,trim=2cm 1.5cm 3cm 1cm]{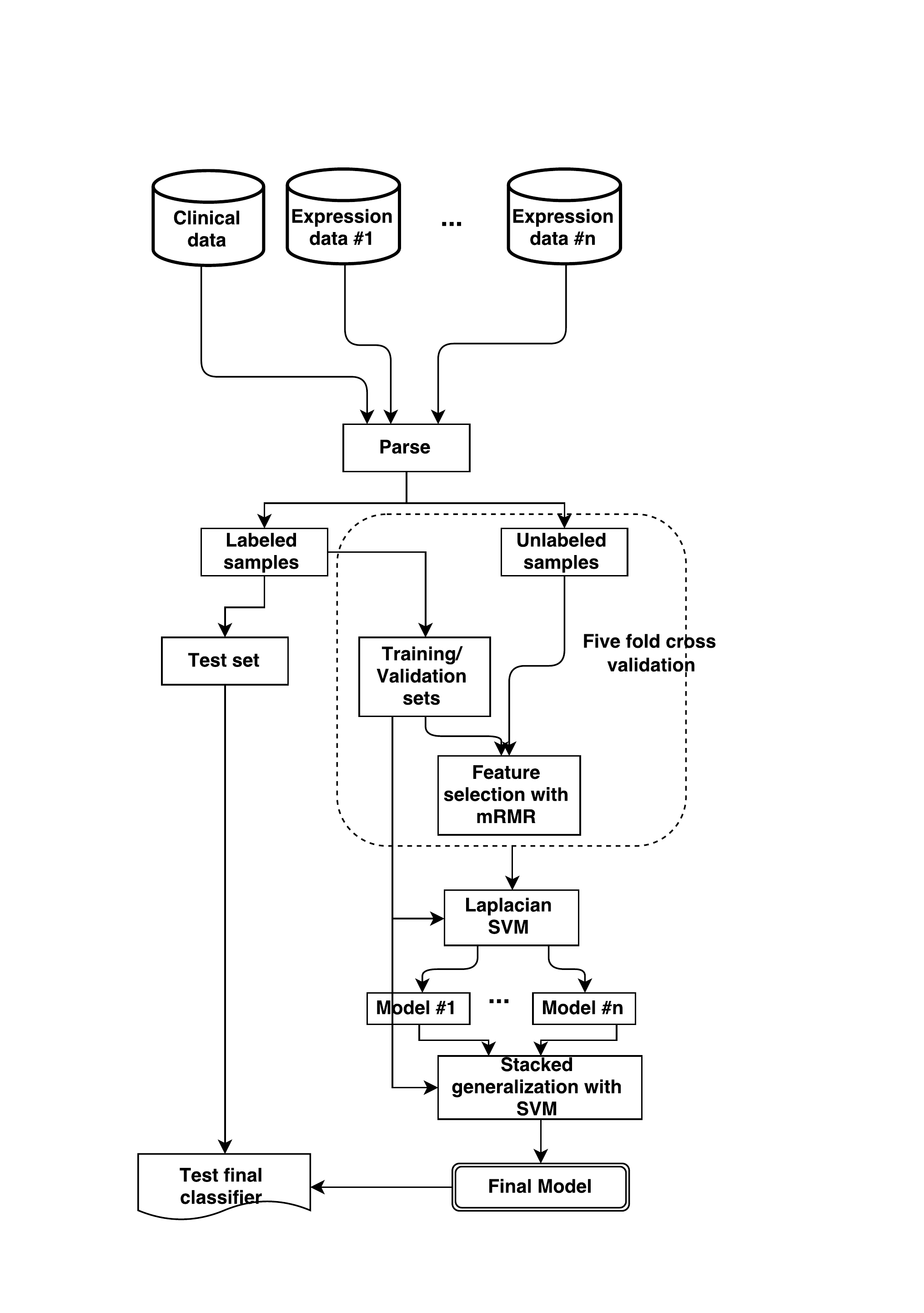}
}
\caption{Pipeline for Semi-Supervised Multi-Modal Prediction Modeling}
\label{fig1}
\end{figure}

Figure \ref{fig1} depicts the block diagram of the proposed pipeline. As illustrated in the figure, it includes four major steps, namely, data pre-processing, feature selection, learning individual SSL models, and finally consolidating these models to produce a combined classifier. 

In the first step, the genomic input data are parsed and stored into feature matrices and survival times are retrieved from the clinical records. Since we are dealing with a classification task, the positive class includes those patients who lived above some threshold of interest (e.g., 5 years for the kidney cancer patients) or are still alive and their survival period as of the last follow-up visit passed that threshold. Similarly, the negative class comprises those samples who deceased before the threshold. For future evaluations and training of the pipeline, we randomly select a subset of the labeled samples (15\%) and split the rest to five folds to carry out five-fold cross-validation according to a specified seed that we change for repeated runs of the same experiments.  

In the next step, to avoid overfitting, it is necessary to perform a feature reduction step. We chose the minimum redundancy maximum relevance (mRMR) \cite{paper20} feature selection technique which has been successfully applied in a number of studies \cite{paper20,paper21,paper22,paper23} that deal with high dimensional data from high-throughput experiments. mRMR is an incremental search algorithm that looks for the sub-set of features with highest relevance to the classes and lowest redundancy. Both relevance and redundancy are defined in terms of the mutual information. Specifically, relevance is defined as the average mutual information between the selected feature subset and the classes. Redundancy, on the other hand, is defined as the average pairwise mutual information between all the samples in the subset. Clearly in the ideal case where all the selected features are independent, the pairwise mutual information between any two features is zero and as a result, the redundancy measure will be zero. Since, this is a multi-objective optimization and there could be many non-dominated solutions, a single objective function is defined as a weighted sum of relevance and negative redundancy. As our goal here is to select those features that are the key determinants of the long-term vs the short-term survival, we computed the z-score of each modality input matrix and discretized its entries into three levels corresponding to those feature values that fall within the ranges $(-\infty,1.5\sigma]$, $(-1.5\sigma,1.5\sigma)$ and $[1.5\sigma,\infty)$. Clearly, the set of differentially expressed genes (DEGs) are expected to fall within the two extreme ranges, whereas those genes which are uncorrelated to the clinical endpoint, are assumed to exhibit smaller variability across the positive and the negative classes. To find the optimal size of selected feature sets as well as the pipeline hyper-parameters (e.g. ambient and intrinsic space regularizers,$\gamma_I$ and $\gamma_A$) we used the validation set and conducted a grid search to pick the best configurations. 

Once the number of features have been reduced, we train multiple Laplacian SVM classifiers, one for each modality. We used the library developed by the authors in \cite{paper16} to train our models. The heat (Gaussian) affinity kernel was selected to generate the edge weights and Euclidean distance function to compute the distance between the nodes. We used 5 nearest neighbor method to filter out the links between distant nodes. Moreover, we selected polynomial kernel of degree three (cf. equation \ref{eq2}) which turns out to strike a good balance between the generalizability and the expressivity of the individual models. Lastly, to come up with a consolidated model, we adopted the stacked generalization strategy \cite{paper24} to weigh each of the individual models according to their prediction scores. This second layer model is trained over the prediction scores produced by the single modality sub-models using a linear-kernel support vector machine and thereby the prediction scores from the previous layer are normalized so that no single modality model dominates the role of other models just by having a higher range of prediction scores. 

\section{Results}
We used the proposed pipeline to predict survival of the neuroblastoma (NB) and the kidney (KIRC) cancer patients. In this section, we address two critical questions: (1) Does the SSL approach improve prediction performance? And (2) How does the overall pipeline perform relative to the individual data models (denoted by Model \textit{\#1, ..., \#n} in Figure \ref{fig1})? 
\subsection{Supervised vs. semi-supervised strategies}
\begin{figure}[t!]
    \centering
    \begin{subfigure}[t]{0.5\textwidth}
        \centering
        \includegraphics[scale=0.5,trim=4cm 8cm 4cm 8cm]{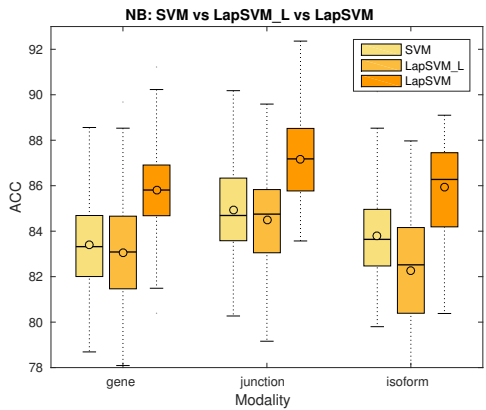}
        \caption{Neuroblastoma}
			\label{fig3}
    \end{subfigure}\\
    \begin{subfigure}[t]{0.5\textwidth}
        \centering
        \includegraphics[scale=.5,trim=4cm 8cm 4cm 8cm]{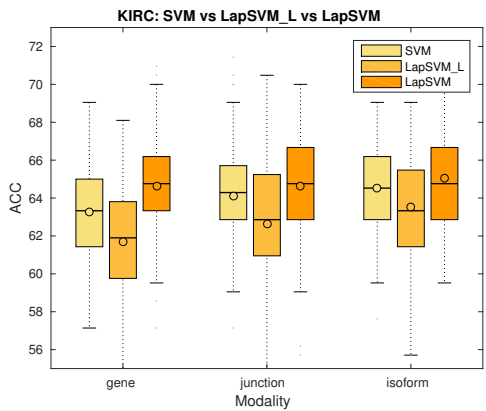}
        \caption{Kidney renal cancer}
			\label{fig2}
	\end{subfigure}%
    \caption{Comparison of SVM and LapSVM }
		\label{fig23}
\end{figure}

We compared the performance of the Laplacian SVM classifier (LapSVM) to its supervised counterpart, the support vector machines (SVM). We evaluated the performance of each model 100 times using different cross-validation training, test, and, validation sets. Figure \ref{fig23} depicts the box plot of the resulting performance measures, for both the neuroblastoma and kidney datasets and the three RNA-seq modalities mentioned above. Interestingly, for the NB dataset, the trained models (both supervised and semi-supervised) can accurately predict the 9-year survival with the semi-supervised models doing significantly better ($p=0.0, 0.0, 0.0$ for gene, junction, and isoform, respectively). For the KIRC dataset, although the LapSVM outperforms the supervised SVM on average, the performance improvement is not statistically significant for the junction and isoform modalities ($p=3.5e-4, 0.2, 0.15$ for gene, junction, and isoform, respectively.) This is because of the more heterogeneous nature of the kidney cancer and the low predictivity of RNA-seq for this dataset, as evidenced by the lower prediction accuracies compared to the neuroblastoma dataset.

To make the role of unlabeled data more tangible, in another series of experiments, we trained the semi-supervised LapSVM models with and without the unlabeled data. In other words, we would like to uncover the impact of adding the unlabeled data on the prediction accuracy. To this end, we repeated the same set of experiments as in the previous subsection but this time with an additional semi-supervised model that does not leverage the unlabeled samples (LapSVM\_L). Figure \ref{fig23} shows the resulting accuracies. According to the figure, adding the unlabeled samples significantly improves the prediction accuracy. In fact when unlabeled samples are not used, the average performance falls below that of the supervised SVM models. 

\subsection{Semi-supervised multi-modal pipeline vs. individual data models}
Finally, we investigated whether the semi-supervised multi-modal pipeline performs at least as best as the individual sub-models. Moreover, we computed similar results for the supervised counterpart of the proposed pipeline, that is, when the semi-supervised based single modality models are replaced with the supervised SVMs. Table \ref{tab3} illustrates the mean along with the standard variation of the prediction accuracies across 100 randomly initialized settings both for the individual models and the pipelines. According to the Table a few key points become evident. First, the semi-supervised models consistently surpass the supervised ones and, second, by using a stacking strategy the prediction accuracy increases synergistically (as is the case for the KIRC dataset) or leads to performances that closely follow the best sub-models (as is the case with the NB dataset.)\\
\begin{table*}[]
\centering
\caption{Single modality vs integrated model: accuracy comparison}
\label{tab3}
\begin{tabular}{  c c c c c }
\hline
Cancer type& Modality & Mean ACC ($\pm \sigma$): Sup & Mean ACC ($\pm \sigma$): Semi-Sup   \\ \hline
\multirow{5}{*}{NB}& Isoform  &  82.25\%($\pm 2.60$)    &\bf{85.98\%} ($\pm 1.96$) \\ \cline{2-4}
                     & Gene &       83.04($\pm 2.44$)\%   &\bf{85.84\%} ($\pm 2.05$)  \\ \cline{2-4}
                   & Junction &     84.50($\pm 2.48$)\%    &\bf{87.16\%} ($\pm 1.87$) \\ \cline{2-4}
                   & Combined &     84.93($\pm 1.79$)\%     &\bf{86.83\%} ($\pm 1.89$) \\ \hline
\multirow{5}{*}{KIRC}& Isoform  &    63.52($\pm 3.10$)\%      &\bf{65.02\%} ($\pm 2.94$) \\ \cline{2-4}
                     & Gene&          61.68($\pm 3.07$)\%      &\bf{64.67\%} ($\pm 2.74$) \\ \cline{2-4}
                   & Junction &       62.65($\pm 3.19$)\%   &\bf{64.68\%} ($\pm 2.90$) \\ \cline{2-4}
                   & Combined &    66.07\%($\pm 2.54$)      &\bf{66.20\%} ($\pm 2.60$) \\ 
\end{tabular}
\end{table*}
According to the table the combined model predicts the sample labels with higher or competitive accuracies compared to the individual models. 

\section{Discussion}
Prediction of survival is a critical step in decision making for patient's therapy. A wrong estimate on patient's survival may lead to a choice which undermines the quality of life or the success of the selected treatment. The current study aims to help doctors to make more objective estimates based on the genomic data recorded during the course of patient's treatment. In so doing, we developed a pipeline that predicts cancer survival using multiple modalities of high-dimensional transcriptomic data. We showed that by exploiting the manifold structure of the cancer input sources and integrating multiple modalities into a semi-supervised learning framework we can achieve high accuracy. We showed that omics data alone can explain a significant portion of the clinical outcome.

While we designed this pipeline for the task of cancer survival prediction, its application is not limited to this area. In fact, such a pipeline is applicable to other domains where there are not enough labeled data available and plenty of unlabeled data exist. 

In this study, we only focused on transcriptomic data modalities. However, a more comprehensive approach may involve integration of different levels of genomic data. In fact, cancer has been known to result from dysregulation by multiple molecular mechanisms \cite{paper25,paper26} which can manifest itself in changes in the DNA structure, copy number, DNA methylation, histone modification, and miRNA regulation. As a result, no single level of genomic data by itself can explain the outcome independently. Therefore, it may be beneficial to integrate the results of other high-throughput experiments in different genomic levels to boost the prediction accuracy which highlights the role of big-data analytics. The integration of different layers of genomic data using SSL has been explored in some of the recent published works such as \cite{paper9,paper15}. Another direction that may be worth following is the integration of both clinical and omics data. Clinical prognostic factors convey valuable information that is not available in omics data. Even if this information exists, we may not be able to capture it or effectively process it in the pre-processing and feature reduction phases. Finding an effective way to combine these two inherently different sources of data may be interesting future research.

\bibliographystyle{abbrv}
\bibliography{Paper}

\end{document}